\documentclass{article}
\usepackage{spconf,amsmath,graphicx}
\usepackage{booktabs}
\usepackage{graphicx, color,xcolor}

\title{Integrating Contrastive Learning into a Multitask Transformer Model for Effective Domain Adaptation}

\twoauthors
 {Chung-Soo Ahn, Jagath C. Rajapakse\sthanks{correspondence to asjagath@ntu.edu.sg}}
	{
	School of Computer Science and Engineering\\
	Nanyang Technological University, Singapore}
 {Rajib Rana}
	{School of Mathematics, Physics and Computing\\
	University of Southern Queensland, Australia}

\begin{document}
%
\maketitle
\begin{abstract}
While speech emotion recognition (SER) research has made significant progress, achieving generalization across various corpora continues to pose a problem. We propose a novel domain adaptation technique that embodies a multitask framework with SER as the primary task, and contrastive learning and information maximisation loss as auxiliary tasks, underpinned by fine-tuning of transformers pre-trained on large language models. Empirical results obtained through experiments on well-established datasets like IEMOCAP and MSP-IMPROV, illustrate that our proposed model achieves state-of-the-art performance in SER within cross-corpus scenarios.
\end{abstract}
\begin{keywords}
Contrastive learning, cross-corpus adaptation, domain adaptation, speech emotion recognition, transformers
\end{keywords}
\section{Introduction}
\label{sec:intro}

Speech Emotion recognition (SER) has been an active research area for decades, but generalization across multiple corpora has not been fully addressed. Existing SER methods perform poorly when tested on data from different sources, especially when dataset sizes are different. This motivates us to propose a generalized approach for domain adaptation for SER, overcoming the limitations of data constraints. 

A substantial amount of progress has been made through pre-trained transformer models in speech recognition and audio representation learning \cite{baevski2020wav2vec}. Moreover, it is verified that information in pre-trained transformers on language models is easily transferable to emotion recognition \cite{yang2021superb}.  In this paper, we intend to tackle cross-corpus SER by exploiting rich knowledge embedded in pre-trained transformers to enhance the generalizability of the models. Here, we address the cross-corpus SER problem where the emotion recognition model is trained on one corpus and tested on a different corpus.

Contrastive learning is one of the recent successful paradigms in self-supervised learning that can learn structures between data when sample sizes are small \cite{oord2018representation}. Its applicability to domain adaptation has started to gain popularity as well \cite{wang2022cross}.
Yet, contrastive learning is under-explored in the cross-corpus SER problem. In particular, how it can be effectively used for domain adaption in SER needs further exploration.

Information Maximization (IM) loss follows cluster assumption \cite{krause2010discriminative}, that a better classifier will learn to find class boundaries to have large margins. IM loss is optimized without labels, simply optimizing entropy computed from only logits. It has strong potential to be used in domain adaptation in SER.


In this paper, we address all the above challenges. The contributions of this paper are as follows:

\begin{enumerate}
\item We propose a multitask framework for domain adaptation in SER, underpinned by a pre-trained transformer, where SER is the primary task and contrastive learning for learning structure of the source and target corpus data are used as secondary task.
\item We add Information Maximization (IM) loss for clustering another secondary task as it can explicitly learn the cluster structure when used together with contrastive learning.
\item We test our model on the cross-corpus experiment, where the model trained on IEMOCAP\cite{busso2008iemocap} and tested on MSP-IMPROV\cite{busso2016msp} and achieve state-of-the-art performance with 10\% improvement.
\end{enumerate}

\section{Related Work}
\label{sec:related}

In this section, we discuss the literature while clustering them into three groups. In the first group, we present the studies that introduce multitask learning for domain adaptation in SER. Semi-supervised learning which makes use of unsupervised learning objectives along with emotion classification objective has been proposed \cite{parthasarathy2020semi}. Such is ladder network that uses layer-wise reconstruction loss that allows usage of additional unlabelled data. Auto-encoder type reconstruction loss was also attempted\cite{neumann2019improving, dissanayake2020speech}. Another alternative that requires reconstruction loss is GAN-based approaches which also is implemented together with synthetic data generation to aid training \cite{su2022unsupervised, latif2022self}. Additional learning objective can be achieved by other labels, such as languages, which relatively easier to obtain and can be used as auxiliary task\cite{latif2022multitask}. 

In the second group, we present studies using pre-trained transformer for cross-corpus SER. Thanks to recent advancement in self-supervised learning with pre-text tasks, such as Wav2Vec2\cite{baevski2020wav2vec}, its knowledge transfer to emotion has become more popular\cite{yang2021superb}. Following the trend, pre-trained transformer was combined with domain adversarial learning \cite{gao2023adversarial} to achieve cross-corpus SER. VGGish transformer with spectrogram input was used to cross-corpus SER \cite{arezzo2022speaker}, where the model was pre-trained on speaker recogition task and its transferred knowledge helped domain adaptation. It is also common to only take resulting embedding from pre-trained transformer as input and build new transformer and train from scratch\cite{zhang2022unsupervised}. 

Summarising the existing studies, we note that none of the existing studies have used a pre-trained transformer within a multitask learning framework while using contrastive learning as a secondary task to improve the accuracy of the primary SER tasks in a cross-corpus setting. This confirms the novelty of our approach in contrast to the existing literature.

\section{Methods}
\label{sec:methodology}
\begin{figure}
    \centering
    \includegraphics[width=1\linewidth]{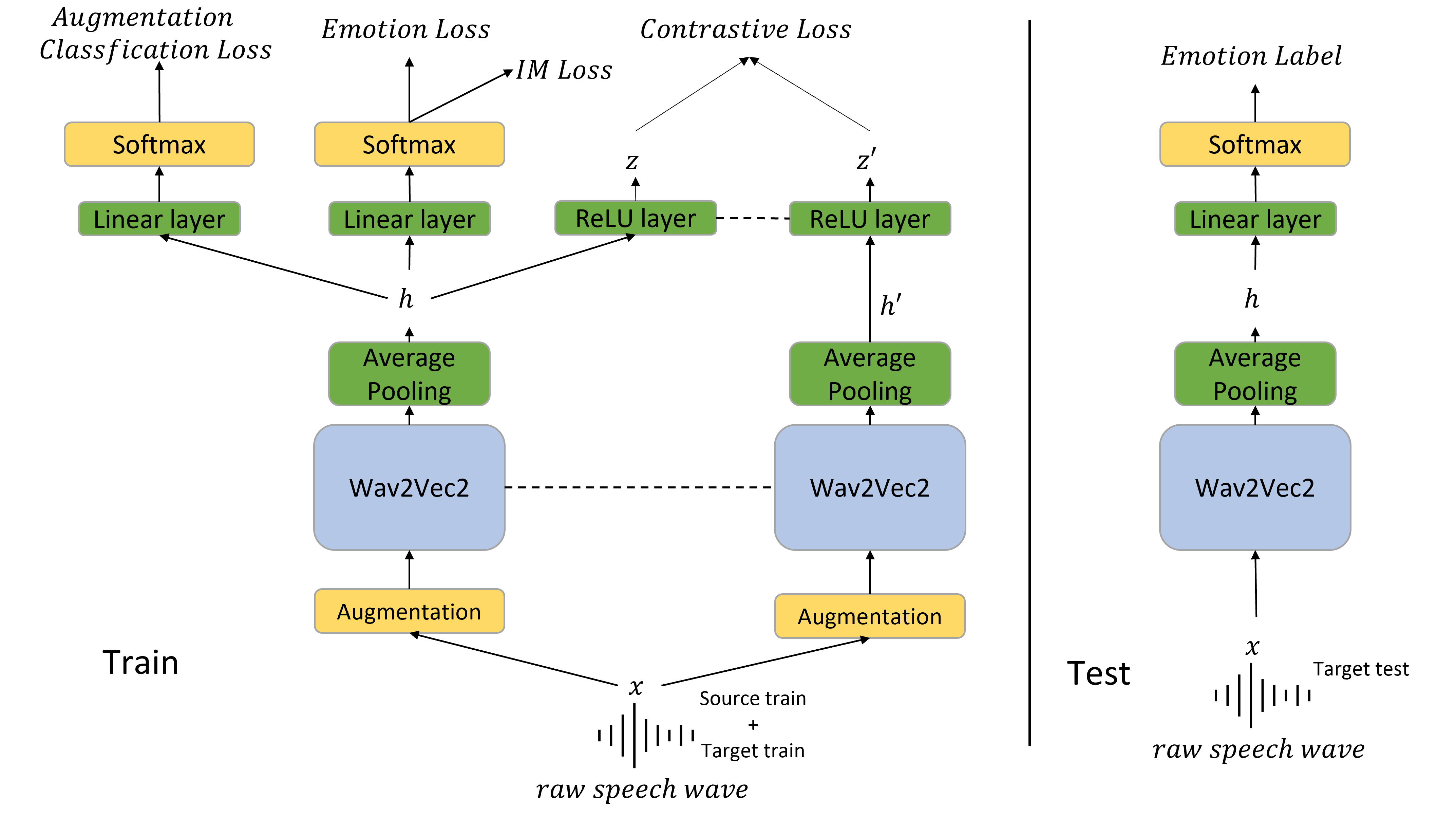}
    \caption{Overall architecture of contrastive learning on speech processed by shared pre-trained transformer.}
    \label{fig:1}
\end{figure}

Wav2Vec2\cite{baevski2020wav2vec} is adopted and two parallel stream is implemented, while sharing same weights. For each speech sample, two augmented speech is generated and fed into those two parallel transformers, as shown in Fig.\ref{fig:1}. Our method also adopts multitask learning framework, that emotion classification layer is on top of transformer and other layers for other auxiliary tasks. Both source data and target data for training will be used to compute loss and train the model through back propagation.

\subsection{Wav2Vec2 for SER}
Wav2Vec2 \cite{baevski2020wav2vec} is self-supervised learning model that takes input from raw waveform data of speech. It is consisted of convolution layers for feature extraction and transformer layers. Transformer layers are consisted of 12 layers of transformer modules. Inspired from Yang et al. \cite{yang2021superb}, we take representations from every transformer layer (total 13 representations) and compute their weighted sum which is learned during training. The output of a transformer is average-pooled and fed to followed by a softmax layer.

The emotion class label $c$ given the weighted output $h$ of the pre-trained transformer is given by 
\begin{align}
p\left(c|h\right)&=\mathrm{softmax}(Wh+b), \label{F01}%
\end{align}
The weights $W$ and biases $b$ of the fully connected layers are learned during training by minimizing the cross-entropy loss $\mathcal{L}_{emo}$:
\begin{align}
  \mathcal{L}_{emo} &= -E\{\sum_{c}{1}_c(y)log(p(c|h))\}. \label{F02}%
\end{align}
where $y$ depicts the target emotion label, ${1}_c$ the indicater function and ${E}$ computes expectation over the samples. .

\subsection{Contrastive learning}

Contrastive learning is aimed at learning to attract positive pairs of inputs and repel negative pairs. Most common and popular variants now uses InfoNCE \cite{oord2018representation} loss to achieve this. We follow the framework from Chen et al.\cite{chen2020simple}, generating latent variable $z$ by  feeding $h$ through single hidden layer feed-forward network, 
\begin{align}
    z=V\sigma(Uh), \label{F03}
\end{align}
where $\sigma$ is ReLU activation. It is known that contrastive learning is more effective with $z$ than $h$\cite{chen2020simple}. With this setting, we compute InfoNCE loss between representations $z$ and $z^\prime$ as follows:

\begin{align}
\mathcal{L}_{cont} &=  \frac{\exp\left(cosine\_sim(z_i,z'_i)/\tau\right)}{ \sum_{j=1}^{batchsize}\exp\left(cosine\_sim(z_i,z'_j)/\tau\right)}. \label{F05}
\end{align}
$z_i$ refers to $i^{th}$ sample of latent variable $z$ in the mini batch and $z'_i$ refers to the $z$ from the forward pass of other augment sample. Hyperparameter $\tau$ is temperature\cite{hinton2015distilling} which controls the smoothness of softmax function. $cosine\_sim$ refers to cosine similarity score \footnote{
$ cosine\_sim(a,b) = \frac{ab } {\left\Vert a \right\Vert \left\Vert b  \right\Vert }$
 }

 Attracting similar data points and repelling dissimlar data points can reveal structure of the dataset. We intend to achieve clustering effect on both source and target data at the same time.
 
\subsection{Information Maximization loss}

Additional auxilliary task is introduced in our work, infomation maximization loss. This is inspired from Krause et al. \cite{krause2010discriminative} and Liang et al.\cite{liang2020we}. Information Maximization (IM) loss follows cluster assumption \cite{krause2010discriminative}, that better classifier will learn to find class boundaries to have large margins. IM loss is optimized without labels, simply optimizing entropy computed from only logits. However, this might lead to trivial solution where every data samples collapse into single class. \cite{krause2010discriminative} proposed uniform distribution contraint for each cluster (or class).

Finally, we compute IM loss by computing expected entropy subtracted by empirical label's entropy. 
\begin{equation} \begin{split}
    \mathcal{L}_{IM}=-{E}\{\sum_{c}p(c|h)log(p(c|h))\} \\+ {E}\{\sum_{c}\hat{p}(c)log(\hat{p}(c))\}. \label{F08}
\end{split}
\end{equation}

where empirical label distribution (simply average of softmax output from all data points) as 
$
\hat{p}\left(c\right)=\frac{1}{N}\sum_{i=1}^{N} p(c_i|h_i). \label{F07}
$

We use IM loss to aid contrastive learning in forming clusters. IM loss explicitly learns to form wide separation margin, thus capable for our needs.

\subsection{Data augmentation and Augmentation Classification loss}

Data augmentation is important component in our model as it increases data size for transformers to learn avoiding overfitting. Also data augmentation is essential for contrastive learning as it requires different views of the same sample to learn similarities between two augmentations. We exploited augmentation tool (https://github.com/asteroid-team/torch-audiomentations) to augment raw audio waveforms. The augmentation function that we adopted are: Gain, PolorityInversion, Shift, TimeInversion, BandStopFilter, PeakNormalization and AddColoredNoise.
All these functions add perturbation to the waveforms and we used mixtures of them to create five pipelines of augmentation.

When implementing data augmentation pipeline, we also assign labels to the data stating which pipeline was this input is perturbed. With this additional label provided, we add another classification task as auxiliary task and compute its loss as:
\begin{align}
   \mathcal{L}_{aug} &= -{E}\{\sum_{{c}_{a}}{1}_{{c}_{a}}(y_a)log(p(c_a|h))\}. \label{F04}%
\end{align}
The label $y_a$ represents from which augmentation pipeline this input came from. This classification layer is separate from emotion classification layer, thus learnable weights  $W'$ and $b'$ are not same with $W$ and $b$.

\subsection{Multitask Learning for domain adaptation}

Finally, all the above mentioned losses are summed up for final loss function for training.
\begin{math}
\mathcal{L}_{total} = \mathcal{L}_{emo} + \lambda_1\mathcal{L}_{aug} + \lambda_2\mathcal{L}_{cont} + \lambda_3\mathcal{L}_{IM} %
\end{math}
where $\lambda_1$,$\lambda_2$, and $\lambda_3$ are hyperparamter constants to control importance of each loss and to be determined empirically.

This loss is computed same regardless of input coming from source or target. Our work is assuming that small fraction of target data with label is available. If label is not available $ \mathcal{L}_{emo}$ can be dropped during training with target data. Detailed procedure of our domain adaptation can be referred to subsection 4.2.

\section{Experiment}
\subsection{Datasets}

In this subsection, we briefly describe the two datasets used for cross-corpus SER experiments. As transformers take raw waveform as inputs, there is no feature extraction.

\begin{enumerate}
\item IEMOCAP \cite{busso2008iemocap}: IEMOCAP dataset consists of 12 hours conversation between two actors. Total 10 actors were recruited to record five sessions. In this work, we focused on binary emotions: neutral, happy, and excited are labelled as positive; and sad and angry as negative. Recordings with other emotion labels are excluded.

\item MSP-IMPROV \cite{busso2016msp}: MSP-IMPROV dataset is constructed similar to IEMOCAP dataset but with 12 actors and six sessions, which has a relatively larger size than IEMOCAP. This dataset only have four emotion labels: neutral, happy, angry and sad. This labels are converted to binary similar to IEMOCAP dataset.
\end{enumerate}

\subsection{Domain adaptation experiment}

\begin{table}[b]
  \caption{Performance comparison with existing methods}
  \label{Table 1}
  \centering
  \begin{tabular}{cc}
    \toprule
    \textbf{Model}                  & \textbf{UAR}    \\
    \midrule
    CNN-BLSTM\cite{gao2022domain}     & $59.52\% $  \\
    CNN-BLSTM+DANN+CenterLoss\cite{gao2022domain}     & $57.26\% $  \\
    CNN-LSTM\cite{gao2023adversarial}     & $55.73\% $  \\
    DoGAT\cite{gao2023adversarial}   & $59.42\% $  \\
    Ours (without labels)         & $59.39\% $  \\
    Ours           & $69.25\% $  \\
    \bottomrule
  \end{tabular}
\end{table}

Our work is aimed at situations where only a small fraction of target data is available with labels to access. IEMOCAP dataset was used as source data and 90\% of the data was used as training set (denoted by $S_{tr}$) and remaining 10\% was reserved as validation (denoted by $S_{va}$) which was used for early stopping. From target dataset MSP-IMPROV, 5\% of data was taken with labels as the training set (denoted by $T_{tr}$) and remaining 95\% was reserved for testing set (denoted by $T_{te}$). And we also experimented with a situation where labels was considered unavailable for all target data. In this case, we set $30\%$ and $70\%$ ratio of the MSP-IMPROV dataset for $T_{tr}$ and $T_{te}$, respectively.

Domain adaptation was tested by training on $S_{tr}$ and $T_{tr}$ for one epoch alternatively. After each epoch, emotion classification accuracy was measured on $S_{va}$. When the accuracy did not improve from previous epoch, training was halted and UAR (Unweighted Average Recall) on $T_{te}$ was recorded.

We empirically set the hyperparameters of the loss: $\lambda_1 = 0.1$, $\lambda_2 = 0.5$ and $\lambda_3 = 0.5$. Pre-trained transformers with name 'facebook/wav2vec2-base-960h' (downloaded from https://huggingface.co/facebook/wav2vec2-base-960h) which was trained in Wav2Vec2 with hidden layer size of $768$; and we set size for $U$ and $V$ as $256$ and $128$, respectively.  Learning rate was $1\times10^{-4}$ and decayed on plateau by factor of $0.1$ with patience of $5$. Validation accuracy was measured for learning rate decay and the epoch (separate from early stopping validation) was set as $100$ batches for source dataset and $25$ batches for target dataset with labels. When target data was without labels, epoch was set as same with the source and $100$ batches were evaluated to decide on learning rate decay.

Our experiment results is presented in Table.\ref{Table 1}. Our method outperforms the state-of-the-art methods by about 10\%. Even more, our model performs on par with state-of-the-art methods when trained on target data without labels. 

\subsection{Ablation atudy}

\begin{table}[b]
  \caption{Ablation Study}
  \label{Table 2}
  \centering
  \begin{tabular}{cc}
    \toprule
    \textbf{Our Model}                  & \textbf{UAR}    \\
    \midrule
       complete model           & $69.25\% $  \\
       without labels         & $59.39\% $  \\
    \midrule
     without $\mathcal{L}_{aug}$            & $59.10\% $  \\
        without $\mathcal{L}_{IM}$ \& $\mathcal{L}_{aug}$          & $58.68\% $  \\
        without $\mathcal{L}_{cont}$ \& $\mathcal{L}_{aug}$            & $57.37\% $  \\
    \bottomrule
  \end{tabular}
\end{table}

Our ablation study focused on cases where no labels are available, to clearly examine generalizability of our model. We also tested contribution of component losses in the cost function. Contribution of $\mathcal{L}_{cont}$ was the largest and $\mathcal{L}_{aug}$ was the least. Especially, $\mathcal{L}_{aug}$ had negligible impact (UAR decreased by 0.2\% when removed).
We believe that $\mathcal{L}_{cont}$ learns to attract representation from similar data samples and repel representation from dissimilar data samples, thus learning the overall structure of the dataset. And $\mathcal{L}_{IM}$ enforces representation to have larger separation margin, resulting in localizing representations from the same cluster in to a dense region. 

\begin{figure}
    \centering
    \includegraphics[width=1\linewidth]{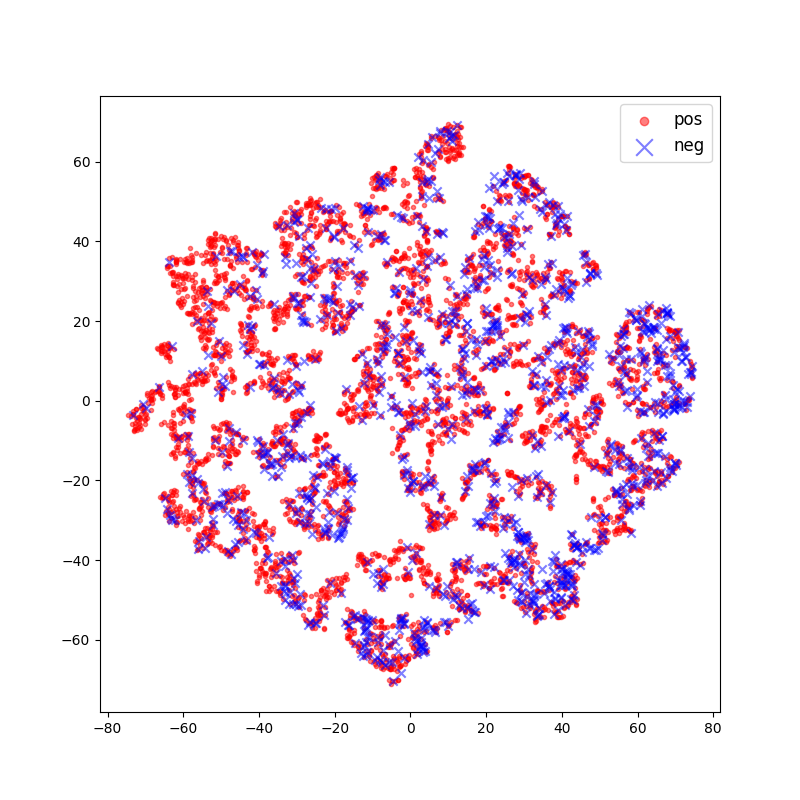}
    \caption{Visualizing $h$ computed from $T_{te}$ from model without $\mathcal{L}_{aug}$.}
    \label{fig:2}
\end{figure}

We also  visualized the data in the latent space in Figure.\ref{fig:2} to test the effect of each loss component. Each marker represents $h$ computed for data from $T_{te}$ without $\mathcal{L}_{aug}$ and differentiated per class. It is notifiable that representations are well dispersed thanks to contrastive learning. The edge region of scatter plot is denser than center and each class is more concentrated either left-end or right-end, thanks to information maximization loss. Finally, contrastive learning and information maximization proves to be effective in domain adaptation as it did not exploit labels of target data to achieve this.

\section{Conclusion}

We proposed to incorporate contrastive learning via multitask learning for domain adaptation in SER. Earlier, contrastive learning was rarely studied on domain adaption problem in context of SER. We proposed to train a transformer model with contrastive learning and other tasks as auxiliary tasks on top of emotion classification. Our experiments demonstrated that multitask learning with contrastive learning was able to learn the structure from data without labels by attracting similar and repelling dissimilar data. Furthermore, information maximization turned out to be capable of pushing representations of different classes to the edge region of different directions for separation. Our experiments showed that our model is able to achieve on par with state-of-the-art in domain adaption without labels. Moreover, our model improved 10\% above state-of-the-art with only 5\% of the target dataset with labels.

\vfill\pagebreak

\bibliographystyle{IEEEbib}
\bibliography{strings,refs}

\end{document}